\documentclass[10pt,twocolumn,letterpaper]{article}

\usepackage[pagenumbers]{cvpr} % To force page numbers, e.g. for an arXiv version

\usepackage{times}
\usepackage{graphicx}
\usepackage{epsfig}
\usepackage{amsmath}
\usepackage{amssymb}
\usepackage{booktabs}

\usepackage{bm}

\DeclareMathOperator*{\argmin}{arg\,min}

\newcommand{\thetab}{\bm{\theta}}
\newcommand{\betab}{\bm{\beta}}

\newcommand{\pp}{\mathbf{p}}

\newcommand{\rr}{\mathbf{r}}

\newcommand{\qq}{\mathbf{q}}

\newcommand{\FF}{\mathbf{F}}
\newcommand{\AB}{\mathbf{A}}
\newcommand{\DD}{\mathbf{D}}

\newcommand{\TT}{\mathbf{T}}
\newcommand{\RR}{\mathbf{R}}
\newcommand{\RRp}{\mathbf{R}^{\prime}}

\newcommand{\MM}{\mathbf{M}}

\newcommand{\Ft}{\mathbf{F_{t}}}

\newcommand{\abs}{\mathbf{a}_{k}}

\newcommand{\fbs}{\mathbf{f}_{k}}
\newcommand{\vbs}{\mathbf{v}_{k}}
\newcommand{\vbstilde}{\widetilde{\mathbf{v}}_{k}}
\newcommand{\vbpstilde}{\widetilde{\mathbf{v}}_{k}^{\prime}}
\newcommand{\VVstilde}{\widetilde{\mathbf{V}}_{s}}
\newcommand{\VVpstilde}{\widetilde{\mathbf{V}}_{s}^{\prime}}

\newcommand{\VV}{\mathbf{V}}
\newcommand{\JJ}{\mathbf{J}}

\usepackage{xspace}

\usepackage[pagebackref,breaklinks,colorlinks]{hyperref}

\usepackage[capitalize]{cleveref}
\crefname{section}{Sec.}{Secs.}
\Crefname{section}{Section}{Sections}
\Crefname{table}{Table}{Tables}
\crefname{table}{Tab.}{Tabs.}

\def\cvprPaperID{8459} % *** Enter the CVPR Paper ID here
\def\confName{CVPR}
\def\confYear{2022}

\begin{document}

\title{HSPACE: Synthetic Parametric Humans Animated in Complex Environments}

\author{
Eduard Gabriel Bazavan\quad Andrei Zanfir\quad Mihai Zanfir \\ William T. Freeman \quad Rahul Sukthankar \quad Cristian Sminchisescu\\
\and
{\bf Google Research}\\
{\tt\small \{egbazavan, andreiz, mihaiz, wfreeman, sukthankar, sminchisescu\}@google.com}\\
}
% \maketitle

\twocolumn[{%
\renewcommand\twocolumn[1][]{#1}%
\maketitle
\begin{center}
    \centering
    \includegraphics[width=0.85\linewidth]{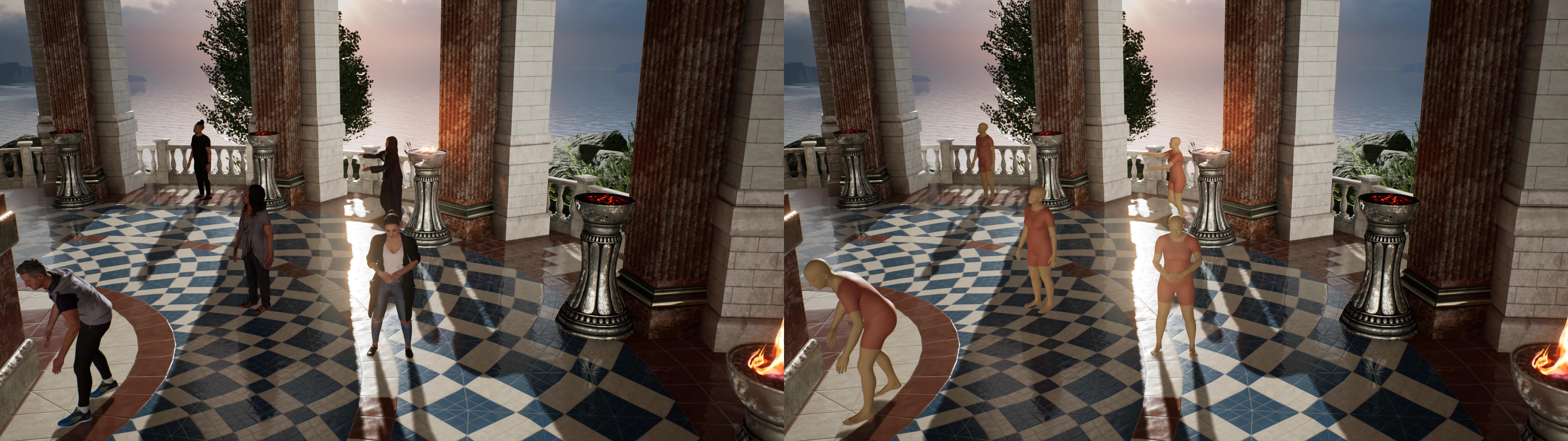}
    \captionof{figure} {\small HSPACE contains dynamic scenes with multiple moving people, with diverse body shapes and poses, placed in realistic environments, under complex lighting. Human animations are driven by GHUM\cite{ghum2020}. For all frames we provide 3d pose and shape ground truth, as well as other rich image annotations including human segmentation, body part localisation semantics, and temporal correspondences.}
    \label{fig:teaser_figure}
\end{center}
}]

%%%%%%%%% ABSTRACT
\begin{abstract}
   Advances in the state of the art for 3d human sensing 
 %human pose, shape, motion estimation, and beyond, \eg inference and reconstruction of clothing, 
   are currently limited by the lack of visual datasets with 3d ground truth, including multiple people, in motion, operating in real-world environments, with complex illumination or occlusion, and potentially observed by a moving camera. Sophisticated scene understanding would require estimating human pose and shape as well as gestures, towards representations that ultimately combine useful metric and behavioral signals with free-viewpoint photo-realistic visualisation capabilities. To sustain progress, we build a large-scale photo-realistic dataset, Human-SPACE (HSPACE), of animated humans placed in complex synthetic indoor and outdoor environments. We combine a hundred diverse individuals of varying ages, gender, proportions, and ethnicity, with hundreds of motions and scenes, as well as parametric variations in body shape (for a total of 1,600 different humans), in order to generate an initial dataset of over 1 million frames. Human animations are obtained by fitting an expressive human body model, GHUM, to single scans of people, followed by novel re-targeting and positioning procedures that support the realistic animation of dressed humans, statistical variation of body proportions, and jointly consistent scene placement of multiple moving people. Assets are generated automatically, at scale, and are compatible with existing real time rendering and game engines. The dataset\footnote{\url{https://github.com/google-research/google-research/tree/master/hspace}} with evaluation server will be made available for research. Our large-scale analysis of the impact of synthetic data, in connection with real data and weak supervision, underlines the considerable potential for continuing quality improvements and limiting the sim-to-real gap, in this practical setting, in connection with increased model capacity.
\end{abstract}

%%%%%%%%% BODY TEXT
\section{Introduction}

Progress in 3d human pose and shape estimation has been sustained over the past years as several statistical 3d human body models\cite{ghum2020,pavlakoscvpr2019,SMPL2015}, as well as learning and inference techniques have been developed\cite{sminchisescu_ijrr03,bogo2016,SMPL2015,ghum2020,dmhs_cvpr17,zanfir2018monocular,Rhodin_2018_ECCV,Kanazawa2018,kolotouros2019learning,ExPose:2020,zanfir2020neural}. More recently there has been interest in human interactions, self-contact \cite{Mueller:CVPR:2021, fieraru2021learning}, and human-object interactions, as well as in frameworks to jointly reconstruct multiple people\cite{jiang2020coherent, zhang2021body, zanfir2018monocular, fieraru2021remips}. As errors steadily decreased on some of the standard 3d estimation benchmarks like Human3.6M\cite{Ionescu14pami} and HumanEva\cite{sigal2010humaneva}, other laboratory benchmarks recently appeared \cite{Fieraru_2020_CVPR,fieraru2021learning}, more complex in terms of motion, occlusion and scenarios, \eg capturing interactions. 

While good quality laboratory benchmarks remain essential to monitor and track progress, as a rich source of motion data to construct pose and dynamic priors, or to initially bootstrap models trained on more complex imagery, overall there is an increasing need to bridge the gap between the inevitably limited subject, clothing, and scene diversity of the lab, and the complexity of the real world. It is also desirable to go beyond skeletons and 3d markers towards more holistic models of humans with estimates of shape, clothing, or gestures. While several recent self-supervised and weakly-supervised techniques emerged, with promising results in training with complex real-world image data\cite{zanfir2020weakly,joo2020exemplar,kolotouros2019learning}, their quantitative evaluation still is a challenge, as accurate 3d ground truth is currently very difficult to capture outside the lab. This either pushes quantitative assessment back to the lab, or makes it dominantly qualitative and inherently subjective. It is also difficult to design visual capture scenarios systematically in order to improve performance, based on identified failure modes.

HSPACE (Synthetic Parametric Humans Animated in Complex Environments) is a large-scale dataset that contains high resolution images and video of multiple people together with their ground truth 3d representation based on GHUM -- a state-of-the art full-body, expressive statistical pose and shape model. HSPACE contains multiple people in diverse poses and motions (including hand gestures), at different scene positions and scales, and with different body shapes, ethnicity, gender and age. People are placed in synthetic complex scenes and under natural or artificial illumination, as simulated by accurate light transport algorithms. The dataset also features occlusion due to other people, objects, or the environment, and camera motion. 

In order to produce HSPACE, we rely a corpus of 100 diverse 3d human scans (purchased from RenderPeople\cite{Renderpeople.com}), with parametric varying shape, animated with over a 100 real human motion capture snippets (from the CMU human motion capture dataset), and placed in 100 different synthetic indoor and outdoor environments, available from various free and commercial sources. We automatically animate the static human scans and we consistently place multiple people and motions, sampled from our asset database, into various scenes. We then render the resulting scenes for different cinematic camera viewpoints at 4K/HDR, using a realistic, high-quality game-engine.

Our contribution is the construction of a large scale automatic system, which requires considerable time as well as human and material resources in order to perfect. The system supports our construction of a 3d dataset, HSPACE, unique in its large-scale, complexity and diversity, as well as accuracy and ground-truth granularity. Such features complement and considerably extend the current dataset portfolio of our research community, being essential for progress in the field. To make the approach practical and scalable we also develop: (1) procedures to fit GHUM to complex 3d human scans of dressed people with capacity to retarget and animate both the body and clothing, automatically, with realistic results, and (2) automatic 3d scene placement methodology to temporally avoid collisions between people, as well as between people and the environment. Finally, we present large-scale studies revealing insight into practical uses of synthetic data, the importance of using weakly-supervised real data in bridging the sim-to-real gap, and the potential for improvement as model capacity increases. The dataset and an evaluation server will be made available for research and performance evaluation.

\section{Related Work}
    There are quite a few people datasets with various degrees of supervision: 2d joint
annotations, semantic segmentations \cite{MsCOCO, OpenImages}, or 3d by fitting a statistical body model or from multi-camera views\cite{zhang2020object, STRAPS2020BMVC, joo2020exemplar, mehta2017monocular},  dense pose \cite{Guler2018DensePose}, indoor mocap datasets with 3d pose ground truth for single or multiple people \cite{sigal2010humaneva, Ionescu14pami, Fieraru_2020_CVPR, fieraru2020three, fieraru2021learning, fieraru2021aifit}, in the wild datasets where IMUs and mobile devices were used to recover 3d pseudo ground truth joints \cite{vonMarcard2018}. All these datasets contain real images, however the variability of the scenes and the humans is limited and the 3d ground truth accuracy is subject to annotators bias, joint positioning errors (for mocap) or IMUs sensor data optimization errors. It is also difficult to increase the diversity of a real dataset, as one cannot capture the same exact sequence from e.g. a different camera viewpoint. 

In order to address the above-mentioned issues, efforts have been made to generate data synthetically using photorealistic 3d assets (scenes, characters, motions). Some synthetic datasets compose statistical body meshes or 3d human scans with realistic human textures on top of random background images, HDRI backdrops or 3d scenes with limited variability \cite{varol2017learning, yan2021ultrapose, Patel:CVPR:2021, zhu2020simpose}, or rely on game engine simulations to recover human motions and trajectories \cite{caoHMP2020}. Table \ref{tbl:datasets_comparison} reviews some of the most popular datasets along several important diversity axes.
Our HSPACE dataset addresses some of the limitations in the state of the art by diversifying over people, poses, motions and scenes, all within a realistic rendering environment and by providing a rich set of 2d and 3d annotations. 

\begin{table*}[!htbp]
\setlength{\tabcolsep}{0.8em} 
\begin{center}
\scalebox{0.78}{
\begin{tabular}{lllllllllll}
% \begin{tabular}{ccccccccccc}
Dataset            & \#Frames & \#Views & \#Subj.          & \#Motions & Complexity                                        & Image      & GT format           \\ \hline\hline
HumanEva \cite{sigal2010humaneva}          &    $\approx 40k$      &     4/7     & 4                 &      6      & 1 subject, no occlusion                           & lab        & 3DJ  \\
Human3.6m  \cite{Ionescu14pami}        &    $\approx 3,6M$      &     4     & 11                &       15     & 1 subject, minor occlusion                        & lab        & 3DJ, GHUM/L  \\ 
CHI3D \cite{fieraru2020three}             &     $\approx 728k$     &     4     &       6            &        120    &         multiple interacting subjects, lab                                          &     lab       &           3DJ, GHUM/L. CS         \\
HumanSC3D \cite{fieraru2021learning}         &      $\approx 1.3M$    &     4     &         6          & 172           &     1 subject, frequent self-contact                                              &     lab       &            3DJ, GHUM/L, CS         \\
Fit3D \cite{fieraru2021aifit}             &     $\approx 3M$      &      4    &        13           &  47  &     1 subject, extreme poses                    &      lab      &        3DJ, GHUM/L             \\
TotalCapture \cite{trumble2017total}       &    $\approx 1.9M$      &      8    & 5                 &       10     & 1 subject, no occlusion                           & lab        & 3DJ  \\
PanopticStudio \cite{Joo_2015_ICCV}     &     $\approx1.5M$     &  $480$        & $\approx100$        & $\approx 120$            & multiple subjects, furniture                      & lab        & 3DJ  \\
HUMBI \cite{yu2020humbi}             &    $\approx 300K$      &   $107$       & 772               &       772     & 1 subject, no occlusion                           & lab        & meshes, SMPL        \\
3DPW \cite{vonMarcard2018}              &    $\approx 51k$      &      $1$    & 18                &      $60$      & multiple subjects in the wild                     & natural    & SMPL                \\
MuPoTS-3D \cite{singleshotmultiperson2018}         &     $\approx 8k$     &     $1$     & 8                 &       $\approx 50$     & multiple subjects in the wild                     & natural    & 3DJ  \\
EFT \cite{joo2020exemplar}               &     $\approx 120K$     &       1   & \textgreater 1000 &    0        & multiple subjects, in the wild                    & natural    & SMPL                \\
STRAPS \cite{STRAPS2020BMVC}      &    331      &   1    & 62                &     0       & 1 subject, in the wild                            & natural    & SMPL                \\ \hline
MPI-INF-3DHP-Train \cite{mehta2017monocular} &    $\approx 1.3M$       &        $14$     & 8                 &       8     & 1 subject, minor occlusion                        & composite & 3DJ  \\
SURREAL \cite{varol2017learning}           &    $6.5M$      &       1   & 145               &      \textgreater{}2000      & 1 subject, no occlusion                           & composite & SMPL                \\
3DPeople \cite{pumarola20193dpeople}           &     2.5M     &   4  & 80                &      70      & 1 subject, no occlusion                           & composite & 3DJ  \\
UltraPose\cite{yan2021ultrapose}              &   $\approx 500k$       &     1     & \textgreater{}1000 & 0 & 1 subject, minor occlusions & composite & DeepDaz, DensePose \\
AGORA\cite{Patel:CVPR:2021}              &   $\approx 14k$       &  1        & \textgreater{}350 &      0      & multiple subjects, occlusion & realistic & SMPL-X, masks \\
\hline
\textbf{HSPACE}            & $1M$          &  5 (var)        &    100$\times$16               & 100            & multiple subjects, occlusion                                                  &  realistic          & GHUM/L, masks \\
\hline
\end{tabular}
}
\end{center}
\vspace{-5mm}
\caption{\small Comparison of different human sensing datasets. From left to right columns represent dataset name, number of different frames, average number of views for each frame, number of different subjects, number of motions, the complexity of the scenes, whether the images are captured in indoor lab environments, in the wild natural scenes or are a composite of synthetic and natural images, as well as the type of ground truth offered e.g. 3d joints, type of statistical body mode (SMPL or GHUM), or 3d surface contact signatures (CS).}
\label{tbl:datasets_comparison}
\end{table*}

\section{Methodology}

Our methodology consists of (1) procedures to fit the GHUM body model to a dressed human scan, as well as realistically repose and reshape it (repose and reshape logic), and (2) methods to place multiple moving (dressed) scans animated using GHUM, into a scene in a way that is physically consistent so that people do not collide with each other and with the environment (dynamic placement logic).

\noindent{\bf Statistical GHUM Body Model.} We rely on GHUM \cite{ghum2020}, a recently introduced statistical body model in order to represent and animate the human scans in the scene. The shape space $\boldsymbol{\beta}$ of the model is represented by a variational auto-encoder.
The pose space $\thetab = \left( \boldsymbol{\theta}_{b}, \boldsymbol{\theta}_{lh}, \boldsymbol{\theta}_{rh} \right)$ is represented using normalizing flows \cite{zanfir2020weakly} with separate components for global rotation $\mathbf{r} \in \mathbb{R}^{6}$  \cite{zhou2018continuity} and translation $\mathbf{t} \in \mathbb{R}^{3}$. The output of the model is a mesh $\mathbf{M}\left(\thetab, \betab\right) =  \left(\mathbf{V}, \mathbf{F}\right)$, where $\mathbf{V} \in \mathbb{R}^{10,168 \times 3}$ are the vertices and $\mathbf{F}$ are the $20,332$ faces.

\subsection{Fitting GHUM to Clothed Human Scans}
\begin{figure*}[!htbp]
\begin{center}
    \includegraphics[width=0.78\linewidth]{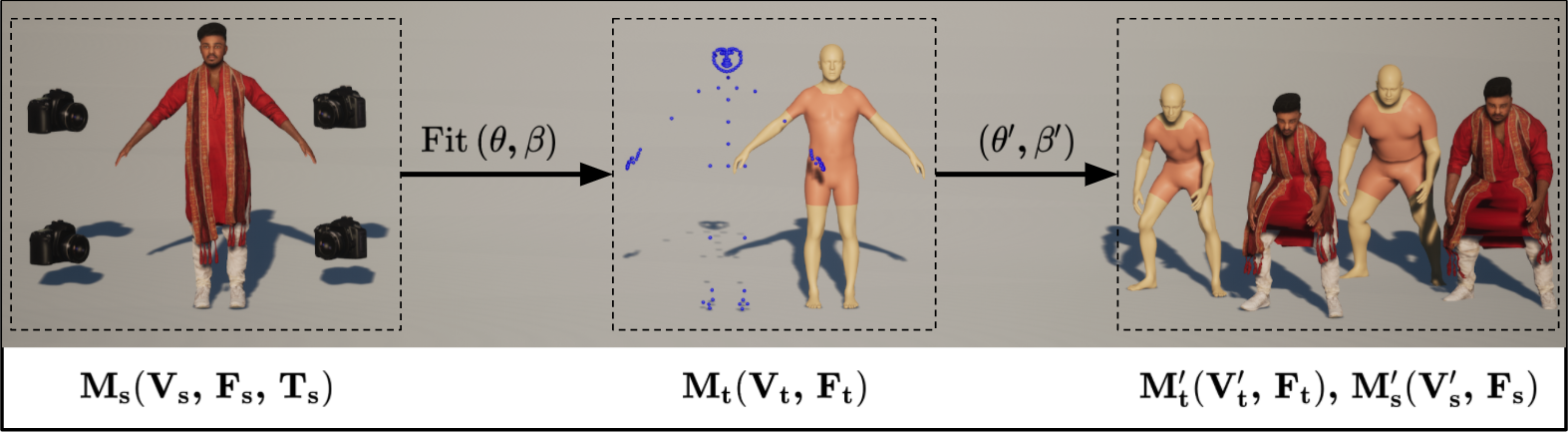}
\end{center}
\vspace{-4mm}
\caption{\small Main processing pipeline for our synthetic human animations. Given a single 3d scan of a dressed person, we automatically fit GHUM to the scan, and build a representation that supports the plausible animation of both the body and the clothing based on different 3d human motion capture signals. Shape can be varied too -- notice also plausible positioning for the fringes of the long blouse outfit.}
\label{fig:fitting_pipeline}
\end{figure*}

The first stage in our pipeline is to fit the GHUM\cite{ghum2020} model to an initial 3d scan of a person $\mathcal{M}_{s} = \left(\VV_s, \FF_s, \TT_s\right)$ containing  vertices $\VV_s \in \mathbb{R}^{N_s}$, faces $\FF_s \in \mathbb{N}^{N_{ts} \times 3}$ and texture information $\TT_s$ containing per vertex $UV$ coordinates and normal, diffuse and specular maps. The task is to find a set of parameters $\left( \thetab, \betab, \rr, \mathbf{t} \right)$ such that the target GHUM\cite{ghum2020} mesh $\mathcal{M}_t\left(\thetab, \betab, \rr, \mathbf{t}\right) =  \left(\VV_t, \FF_t\right)$ is an accurate representation of the underlying geometry of $\mathcal{M}_s$. For the sake of simplicity, we drop the dependence on the parameters $\rr$ and $\mathbf{t}$. As illustrated in fig. \ref{fig:fitting_pipeline}, we uniformly sample camera views around the subject and render it using the texture information associated to $\mathcal{M}_s$. Image keypoints for the body, face, and hands are predicted for each view using a standard regressor \cite{ghum2020,bazarevsky2020blazepose} and we triangulate to obtain a 3d skeleton $\JJ_s \in \mathbb{R}^{N_j \times 3}$ for the source mesh. The fitting procedure of the GHUM mesh $\mathcal{M}_t\left(\thetab, \betab\right)$ to $\mathcal{M}_s$ is formulated as a nonlinear optimization problem with the following objective
\begin{align}
    L\left(\thetab, \betab\right) =& \lambda_{j}L_{j}\left(\JJ_t, \JJ_s\right) + L_{m}\left(\VV_t, \VV_s\right) + \nonumber \\
    & l\left( \boldsymbol{ \theta } \right) + l\left( \boldsymbol{ \beta } \right). \\
    \thetab^{*}, \betab^{*} =& \argmin(L\left(\thetab, \betab\right))
\label{eq:fitting}
\end{align}
In \eqref{eq:fitting}, $\JJ_t \in \mathbb{R}^{N_j}$ are the skeleton joints for the posed mesh $\mathcal{M}_t\left(\thetab, \betab\right)$ and  $L_{j}\left(\mathbf{\JJ_t, \JJ_s}\right) = \frac{1}{N_j} \sum_{i=1}^{N_j}\|\JJ_{s, i} - \JJ_{t, i}\|_2$ is the 3d mean per joint position error between the 3d joints of the source and those of the target. $L_{m}\left(\VV_t, \VV_s\right)$ is an adaptive iterative closest point loss between the target vertices $\VV_t$ and the source vertices $\VV_s$. At each optimization step we split the vertices $\VV_t$ into two disjoint subsets: the vertices $\VV_t^i$ that are inside $\mathcal{M}_s$ and the vertices $\VV_t^o$ which are outside of $\mathcal{M}_s$. In order to classify a vertex as inside or outside, we rely on a fast implementation of the generalized winding number test \cite{Jacobson:WN:2013, fieraru2021remips}. Given the closest distance $d$ between a point $\pp$ and a vertex set $\VV$
\begin{equation}
    c(\pp,\VV)=\min_{\qq \in \VV} d(\pp,\qq)
\end{equation}
we define $L_m$ as follows
\begin{equation}
L_m=\lambda_i\sum_{\pp \in \VV_t^i} c(\pp,\VV_s) + \lambda_o \sum_{\pp \in \VV_t^o} c(\pp,\VV_s)
\label{eq:inside_outside_loss}
\end{equation}

We set $\lambda_i < \lambda_{o}$, enforcing the reconstructed mesh $\MM_t$ to be inside of $\MM_s$, but close to the surface. We add regularization for pose and shape based on their native latent space priors  $l(\thetab)= \|\thetab\|_2^2, \;\;  l(\betab)=\|\betab\|_2^2$ in order to penalize deviations from the mean of their Gaussian distributions.

\subsection{Reposing and Reshaping Clothed People}
\label{sec:reposing}
\begin{figure}[!htbp]
\begin{center}
    \includegraphics[width=0.9\linewidth]{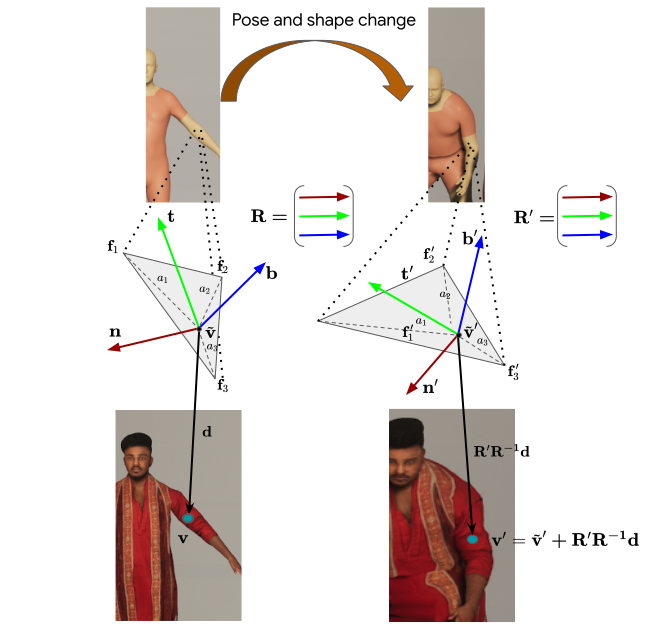}
\end{center}
\vspace{-4mm}
\caption{\small \textbf{Reposing and Reshaping Clothed People.} We compute displacements from GHUM to the scanned mesh in a local coordinate system. For each vertex of the scan, we consider its nearest neighbor point on the GHUM mesh. This point is parameterized by barycentric coordinates. When the GHUM mesh is generated for different pose and shape parameters, its local geometry rotates and scales. We want displacements between the scan and the updated GHUM geometry be preserved. We use a tangent space coordinate system that allows equivariance to rotations. Furthermore, due to the way the tangent space is computed, based on triangle surface area, we are also invariant to scale deformations.}
\label{fig:local_coordinate}
\end{figure}
We design an automated process of generating large-scale animations of the same subject's scan, but with different shape characteristics.
We need the animation process to be compatible with LBS pipelines, such as Unreal Engine, in order to automate the rigging and rendering process for large-scale data creation. This is a non-physical process in the absence of explicit clothing models, but we aim for automation and scalability, rather than perfect simulation fidelity. 
We aim not only for animation diversity, but also for shape diversification. We support transformations in the tangent-space of local surface geometry  that can accommodate changes in both shape and pose – this is different from inverse skinning methods ~\cite{huang2020arch} that only handle the latter.

\vspace{3mm}

\noindent{\bf Tangent-space representation.} Given a source scan with mesh $\MM_s$ and its fitted GHUM mesh $\MM_t$, we compute a displacement field $\DD \in \mathbb{R}^{N_{s} \times 3}$ from $\MM_s$ to $\MM_t$. For each vertex $\vbs \in \VV_s$ we compute its closest point $\vbstilde$ on $\MM_t$ and we denote by $\abs \in \mathbb{R}^{3}$ its barycentric coordinates on the projection face $\fbs \in \FF_t$. From all values $\abs, k \in 1, \ldots, N_{s}$ we build a sparse connection matrix $\AB \in \mathbb{R}^{N_{s}  \times N_{t}}$. The displacement field 
$\DD$ from $\MM_s$ to $\MM_t$ is defined as 
\begin{align}
    \DD = \VV_s - \VVstilde,
\label{eq:displacement}
\end{align}
where $\VVstilde \in \mathbb{R}^{N_s \times 3}$ are all the stacked closest points $\vbstilde$ and $\VVstilde = \AB \VV_t$. 

We want each of the displacement vectors $\mathbf{d}_k$ in $\DD$ to reside in a local coordinate system determined by the supporting local geometry $\{\abs, \fbs \in \Ft\}$. Hence, we compute associated normal $\mathbf{n}_k$, tangent $\mathbf{t}_k$ and bitangent $\mathbf{b}_k$ vectors. The normals and tangents are interpolated given per-vertex information available for the faces $\fbs \in \Ft$, $\abs$. Per-vertex tangents are a function of the UV coordinates. For more details on the usage of UV coordinates to obtain tangents, see \cite{premecz2006iterative}. After Gram–Schmidt orthonormalization of tangents and normals, we derive a rotation matrix $\RR_k  = [\mathbf{t}_k; \mathbf{n}_k; \mathbf{t}_k \times \mathbf{n}_k] \in \mathbb{R}^{3 \times 3}$ representing a local coordinate system for each displacement vector $\mathbf{d}_k$. We stack the rotation matrices for all displacement vectors and construct $\RR \in {\mathbb{R}^{N_s \times 3 \times 3}}$.

\noindent{\bf Controlling shape and pose.} For a target set of pose and shape parameters $\left(\thetab^{\prime}, \betab^{\prime}\right)$ of GHUM, let $\MM_t'\left(\thetab^{\prime}, \betab^{\prime}\right) = \left( \VV_t', \FF_t \right)$ be the new target GHUM posed mesh with vertices $\VV_t'$. The task is to find $\MM_s'\left(\VV_s', \FF_s\right)$ which would correspond to the same change in pose and shape for $\MM_s$. For that, we first compute $\VVpstilde = \AB \VV_t'$. Using $\VVpstilde$ and $\MM_t'$ we get updated local orientations $\RR^{\prime}$ for each $\vbpstilde \in \VVpstilde$ from the normal, tangent and bitangent vectors similarly to $\RR$. Note $\RRp\RR^{-1}$ gives the change of orientation for the supporting faces $\fbs \in \FF_t$ from $\vbstilde$ to $\vbpstilde$. We use them to compute the change in orientation for the displacement field $\DD$ 
\begin{align}
    \VV_s' = \VVpstilde + \RRp\RR^{-1}\DD
\end{align}
and obtain the corresponding mesh $\MM_s'\left(\VV_s', \FF_s\right)$. 

\paragraph{Rendering engine compatibility} Rendering engines use linear blend skinning to display realtime realistic animations, so we cannot incorporate tangent-space transformations to drive the animation. Instead, we use tangent-space transformations to compute a new target rest mesh (this is equivalent to unposing and reshaping), with different body shapes sampled from the latent distribution of the GHUM model, and then continue the animation by LBS. We compute the skinning weights for $\MM_s'$ as $\mathbf{W}_{s}^{\prime} = \AB \mathbf{W}_{t}^{\prime}$, where $\mathbf{W}_{t}^{\prime} \in \mathbb{R}^{N_t \times N_j}$ are the skinning weights for $\MM_t'$. The skeleton animation posing values, skinning matrix $\mathbf{W}_{s}^{\prime}$ and updated rest mesh $\MM_s'$ are sufficient for animation export.

\begin{figure*}[!htbp]
\begin{center}
    \includegraphics[width=0.94\linewidth ]{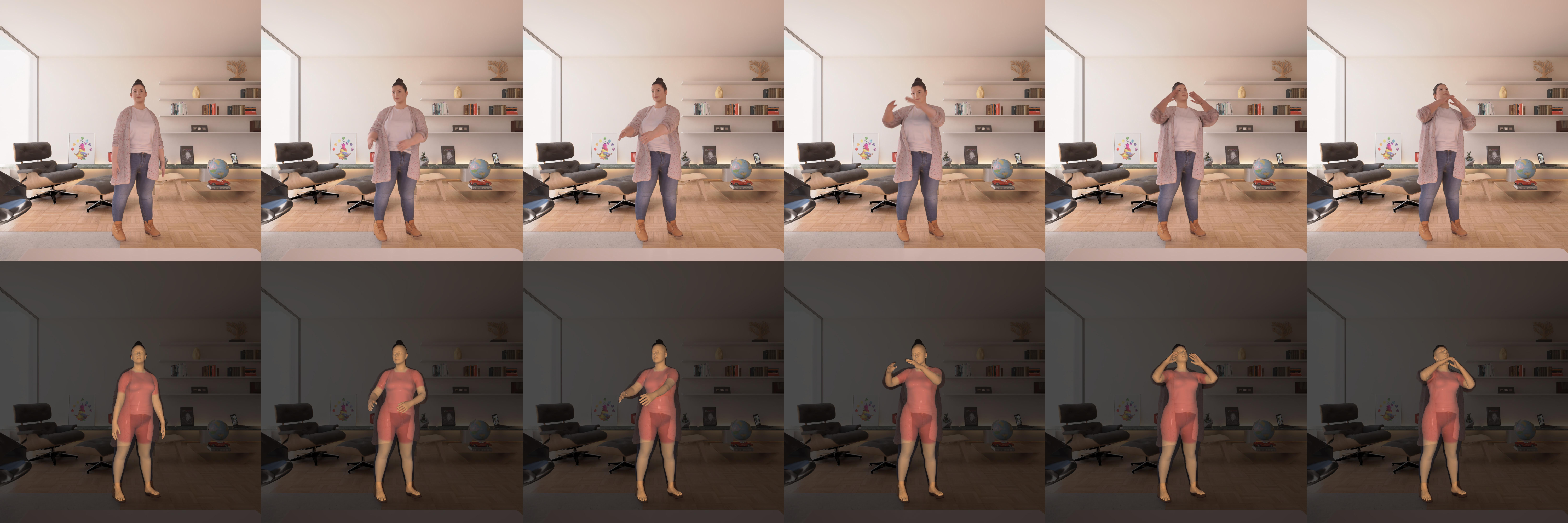}
    \includegraphics[width=0.94\linewidth]{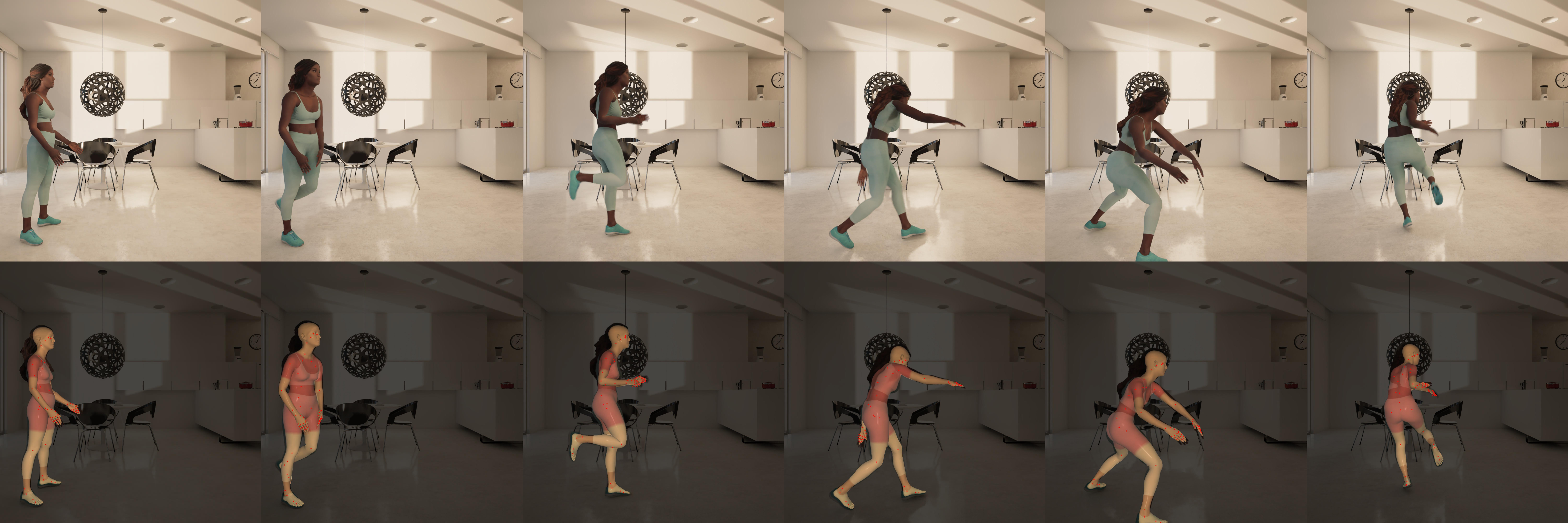}
\end{center}
\vspace{-4mm}
\caption{\small Frames from HSPACE sequences with companion GHUM ground truth. Highly dynamic motions work best with characters wearing tight fitted clothing, the animated sequences look natural and smooth (bottom rows) but also notice good performance for less tight clothing (top rows). See our Sup. Mat. for videos.}
\label{fig:sample_sequences}
\end{figure*}

The limitations of our animation method lie in the hair or clothing simulation which lacks physical realism. However, this geometric animation process is efficient and easy to compute and, as can be seen in fig. \ref{fig:sample_sequences}, results are visually plausible within limits. Our quantitative experiments show that such synthesis methodology improves performance on challenging tasks like 3d pose and shape estimation.

\subsection{Scene Placement Logic}
\label{sec:scene_placement}
In order to introduce multiple animated scans into scenes, we develop a methodology for automatic scene placement based on free space calculations. Typically, we sample several people, their shape, and their motions as well as a bounded, square region of the synthetic scene, so it can be comfortably observed by 4 cameras placed in the corners of the square at different elevations. This is important as some synthetic scenes could be very large, and sampling may generate people spread too far apart or not even visible in any of the virtual cameras. 

The union of tightly bounding parallelepipeds for each human shape at each timestep of their animation defines a 
\emph{motion volume}. These are aligned with a global three-dimensional grid. 
%additionally indexed by the person.
%$g(x,y,z,p)$ associated with the part of the environment being simulated. 
The objective is to estimate a set of positions and planar orientations for the motion volumes, such that no two persons occupy the same unit volume at the same motion timestep (as otherwise trajectories from different people at different timesteps can collide).

Given a scene (3d bounding boxes around any objects including the floor/ground), we sample a set of random motion volumes and initially place them into the scene such that the mid point of the motion paths is in the middle of the scene. We define a loss function which is the sum of {\it a)} number of collision between the sequences (defined as their time-varying 3d bounding boxes intersecting or intersecting with object bounding boxes) and {\it b)} the number of time steps when they are outside the scene bounding box. 
%{\it (c)} Consistency with the 3d motion capture trajectory to follow.

\begin{figure}[!htbp]
\begin{center}
    \includegraphics[width=\linewidth]{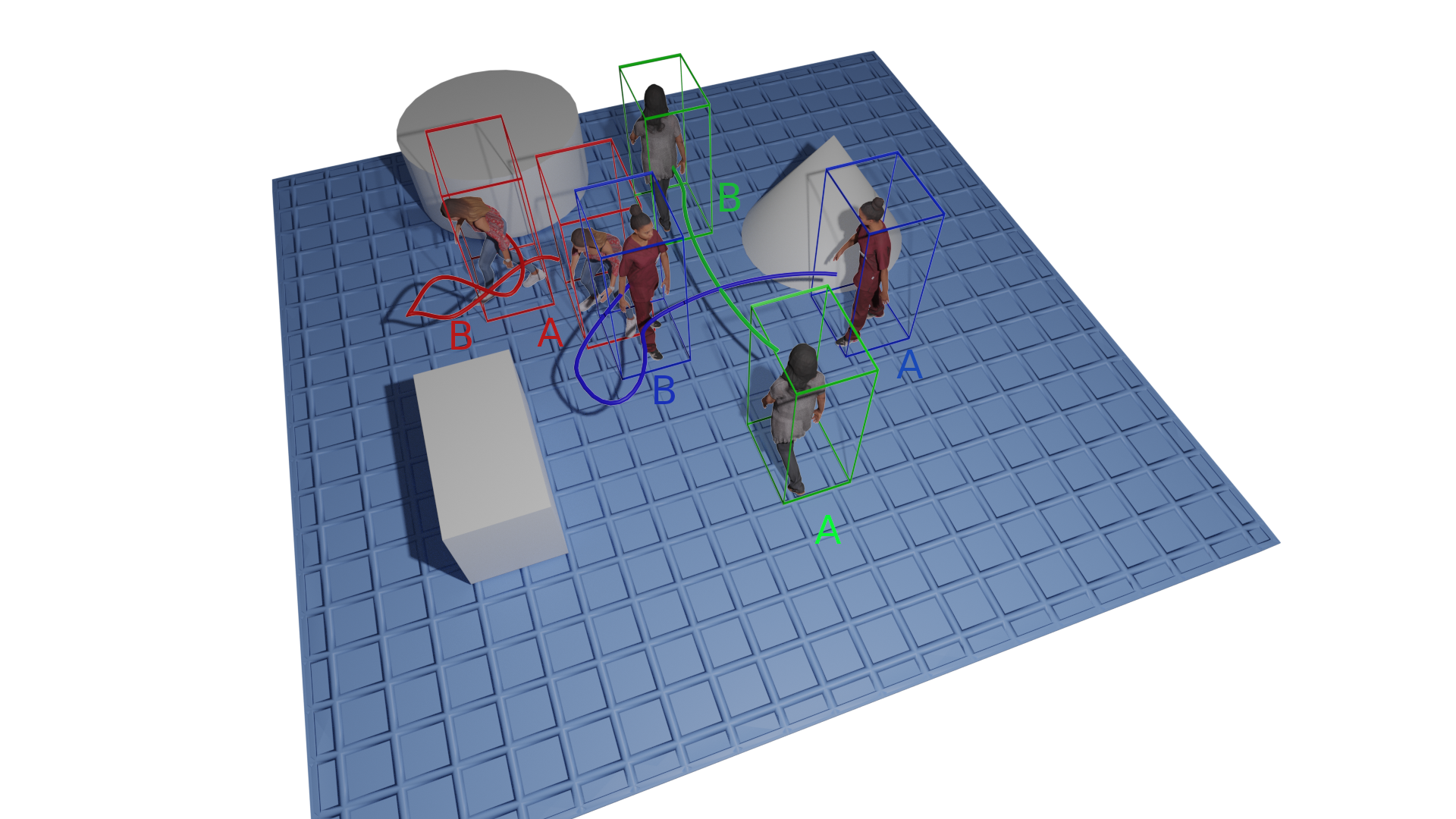}
\end{center}
\vspace{-4mm}
\caption{\small Dynamic placement logic ensures that multiple moving people follow plausible human motions, and are positioned in a scene in way that is consistent with spatial occupancy from other objects or people. An optimization algorithm ensures no two people occupy the same scene location at the same motion timestep. Trajectories are shown in color, with start/end denoted by A/B.}
\label{fig:placement}
\end{figure}

The input to the loss function is a set of per-sequence translation variables, as well as rotations around the axis of ground normal. We then minimize this loss function using a non-differentiable covariance matrix adaptation optimization method (CMA)\cite{Hansen2006} over the initial translation and rotation of the motion volumes, 
%as well as body poses along that volume, 
and only accept solutions where the physical loss is 0 (\ie has no collisions and all sequences are inside the scene bounding box). While the scene placement model can be improved in a number of ways, including the use of physical models or environmental semantics it provides an automation baseline for initial synthesis.
See fig.\ref{fig:placement} for an illustration.

\noindent{\bf Automatic Pipeline.} We designed a pipeline, such that given a query for a specific body scan asset, animation and scene, we produce a high quality rendering placed automatically, at a physically plausible location.

\subsection{HSPACE dataset}

\noindent{\bf Dataset Statistics.} Our proposed HSPACE dataset was created by using $100$ unique photogrammetry scans of people from the commercial dataset RenderedPeople~\cite{renderpeople}. We reshape the scans using our proposed methodology (see section~\ref{sec:reposing}), with $16$ uniformly sampled shape parameters sampled from GHUM's VAE shape space. For animation, we use $100$ CMU motion capture sequences for which we have corresponding GHUM pose parameters. For background variation, we use $100$ complex, good quality 3d scenes. These include both indoor and outdoor environments. To create a sequence in our dataset, we randomly sample from all factors of variation and place the animations in the scene using our scene placement method (see section~\ref{sec:scene_placement}). In total, we collect $1,000,000$ unique rendered frames, each consisting of $5$ subjects on average. An example of a scene with multiple dynamic people is shown in fig.\ref{fig:teaser_figure}.

\noindent{\bf Rendering.} HSPACE images and videos are rendered using Unreal Engine 5 at 4k resolution. The rendering uses ray-tracing, high resolution light mapping, screen-space ambient occlusion, per-category shader models (e.g. Burley subsurface scattering for human skin), temporal anti-aliasing and motion blur. For each frame we capture the ground truth 3d pose of the various people inserted in the scene and save render passes for the finally rendered RGB output, as well as segmentation masks. On average, our system renders at 1 frame/s including saving data on disk. All of our dataset was rendered on 10 virtual machines with GPU support running in the cloud.

\begin{figure*}[!htbp]
\begin{center}
    \includegraphics[width=0.78\linewidth]{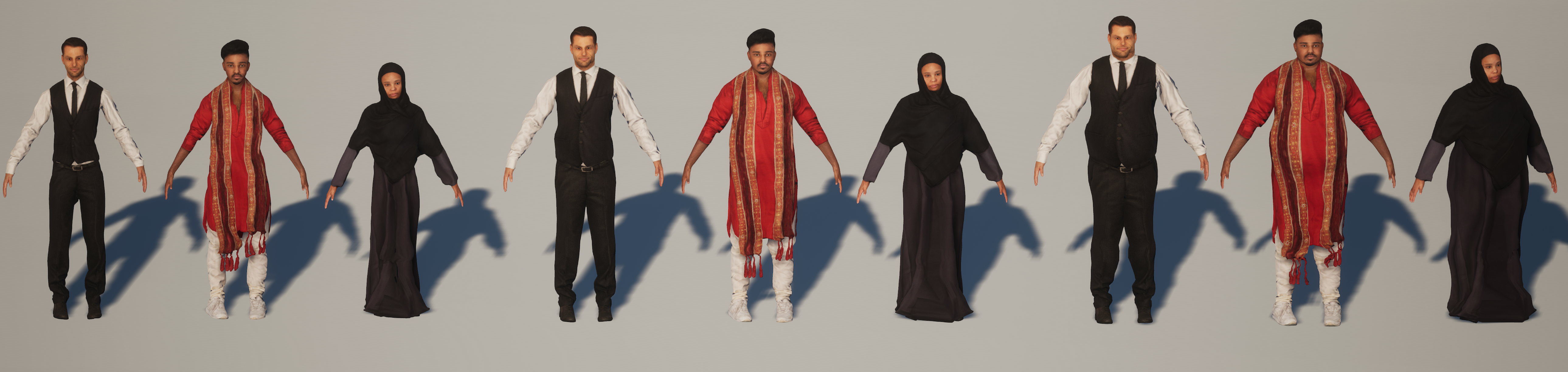}
\end{center}
\vspace{-4mm}
\caption{\small Three scans with different appearance and body mass index, synthesised using GHUM statistical shape parameters, based on a single scan of each subject. Notice plausible body shape variations and reasonable automatic clothing deformation as body mass varies.}
\label{fig:appearance_shape}
\end{figure*}

\begin{figure*}[!htbp]
\begin{center}
    \includegraphics[width=0.96\linewidth]{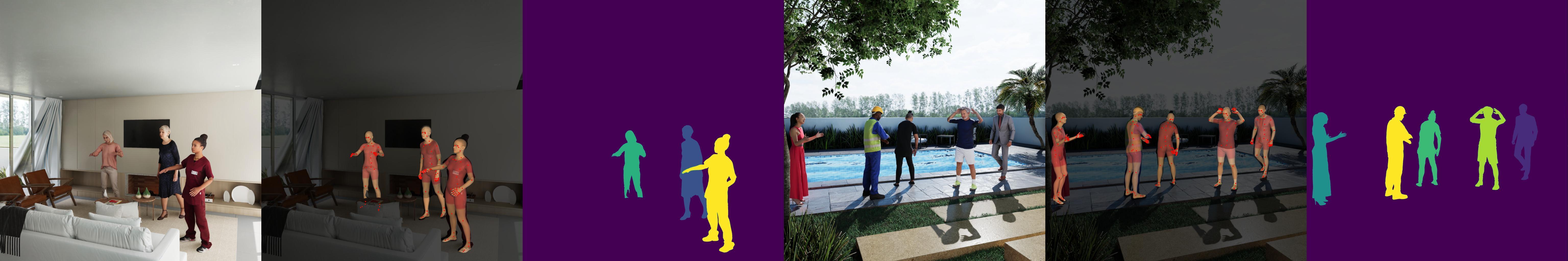}
\end{center}
\vspace{-4mm}
\caption{\small Human scans animated and placed in complexly lit 3d scenes with background motion (e.g. curtains, vegetation).}
\label{fig:sample_render_passes}
\end{figure*}

\section{Experiments}

We validate the utility of HSPACE for both training and evaluation of 3d human pose and shape reconstruction models.
We split HSPACE 80/20$\%$ into training and testing subsets, respectively. We use different people and animation assets for each split.

We additionally employ a dataset with images in-the-wild, \textbf{Human Internet Images (HITI)} (100,000 images), of more than 20,000 different people performing activities with highly complex poses (e.g. yoga, sports, dancing). This dataset was collected in-house and is annotated with both 2d keypoints and body part segmentation. We use it our experiments for training in a weakly supervised regime. The test version of this dataset, \textbf{Human Internet Images (HITI-TEST)}, consists of 40,000 images with fitted GHUM parameters under multiple depth-ordering constraints that we can use as pseudo ground-truth for evaluation in-the-wild (see our Sup. Mat. for details). 

\noindent{\bf Evaluation of GHUM Fitting to Human Scans.} In order to evaluate our GHUM fitting procedure, we compute errors of the nonlinear optimization fit in \eqref{eq:fitting} with keypoints only ($L_{j}$), as well as for the full optimization ($L_{j} + L_{m}$) with $L_{m}$ as well. Results are given in table \ref{tbl:ghum_fitting_errrors}.

\begin{table}[!htbp]
    \small
    \centering
    \begin{tabular}[t]{|l||r|r|}
    \hline
    \textbf{Fitting Method}  & V2V & Chamfer \\
    \hline
    $L_{j}$ & $10$ & $13$ \\
    \hline
    $L_{j} + L_{m}$ & $\textbf{8}$ & $\textbf{11}$ \\
    \hline
    \end{tabular}

    \caption{\small Fitting evaluation with vertex to vertex errors and bidirectional Chamfer distance. Values are reported in mm. Please see fig. \ref{fig:sample_sequences} and our Sup. Mat. for qualitative visualizations.}
\label{tbl:ghum_fitting_errrors}
\end{table}

In all experiments we train models for 3d human pose and shape estimation based on the THUNDR architecture\cite{Zanfir_2021_ICCV}. We report standard 3d reconstruction errors used in the literature: mean per joint position errors with and without Procrustes alignment (MPJPE, MPJE-PA) for the 3d pose and mean per vertex errors with and without Procrustes alignment (MPVPE, MPVPE-PA) for the 3d shape, as well as global translation errors. 

We present the experimental results on the test set of \textbf{HSPACE} in table \ref{tbl:ablation_space}. First we report results for various state of the art 3d pose and shape estimation models such as HUND\cite{zanfir2020neural}, THUNDR\cite{Zanfir_2021_ICCV}, SPIN\cite{kolotouros2019learning} and VIBE\cite{kocabas2019vibe}. The first two methods estimate GHUM mesh parameters, while the last two methods output SMPL mesh parameters. Both SPIN\cite{kolotouros2019learning} and VIBE\cite{kocabas2019vibe} use orthographic projection camera models so we can not report translation errors. We train a  weakly supervised (WS) version  of THUNDR on the HITI training dataset and fine tune it on HSPACE in a fully supervised (FS) regime. This model performs better than all other state of the art methods. The best reconstruction results are obtained by a modified temporal version of THUNDR (labeled as T-THUNDR in table \ref{tbl:ablation_space}) with the same number of parameters as the single-frame version. We provide details of this architecture in the Sup. Mat.

We also train and evaluate on a widely used dataset in the literature, the \textbf{Human3.6M}~\cite{Ionescu14pami} dataset. 
This is an indoor benchmark with ground-truth 3d joints obtained from a motion capture system. 
We report results on protocol P1 (100,000 images) where subjects S1, S5-S8 are used for training, and subjects S9 and S11 are used for testing. In table~\ref{tbl:H36MP1} we show that a refined variant of the THUNDR\cite{Zanfir_2021_ICCV} architecture on HSPACE training data  achieves the lowest reconstruction errors under all metrics. 

We also performed a comprehensive study in order to understand the impact of increasing the size of synthetic data on model performance. Other important factors are the sim-to-real gap, the importance of real data, and the influence of model capacity on performance. One of the most practical approaches would be to use large amounts of supervised synthetic data, as well as potentially large amounts of real images without supervision. The question is whether this combination would help and how would the different factors (synthetic data, real data, model capacity, initialisation and curriculum ordering) play on performance.

\begin{figure*}[!htbp]
\begin{center}
    \includegraphics[width=0.245\linewidth ]{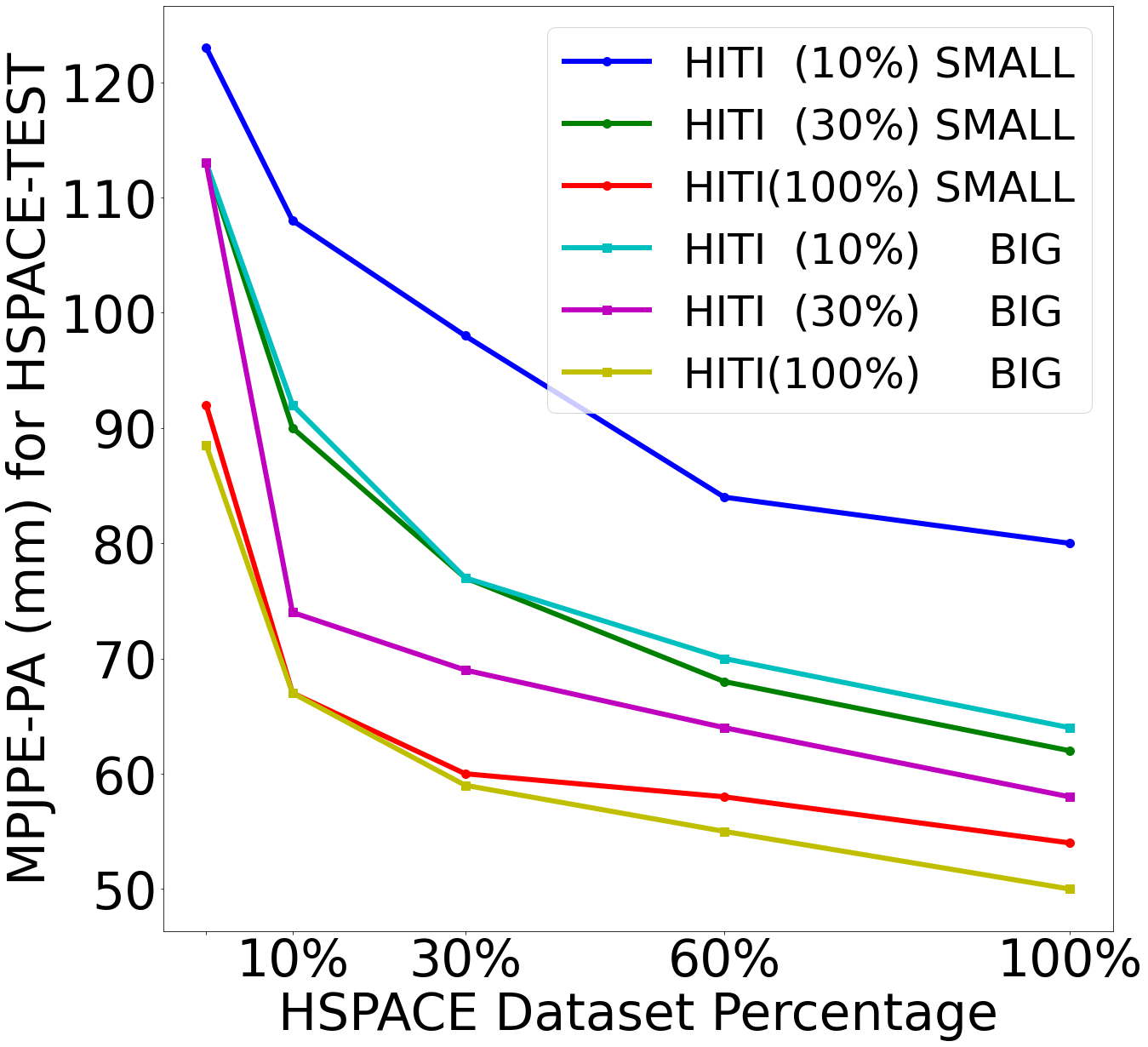}
    \includegraphics[width=0.245\linewidth ]{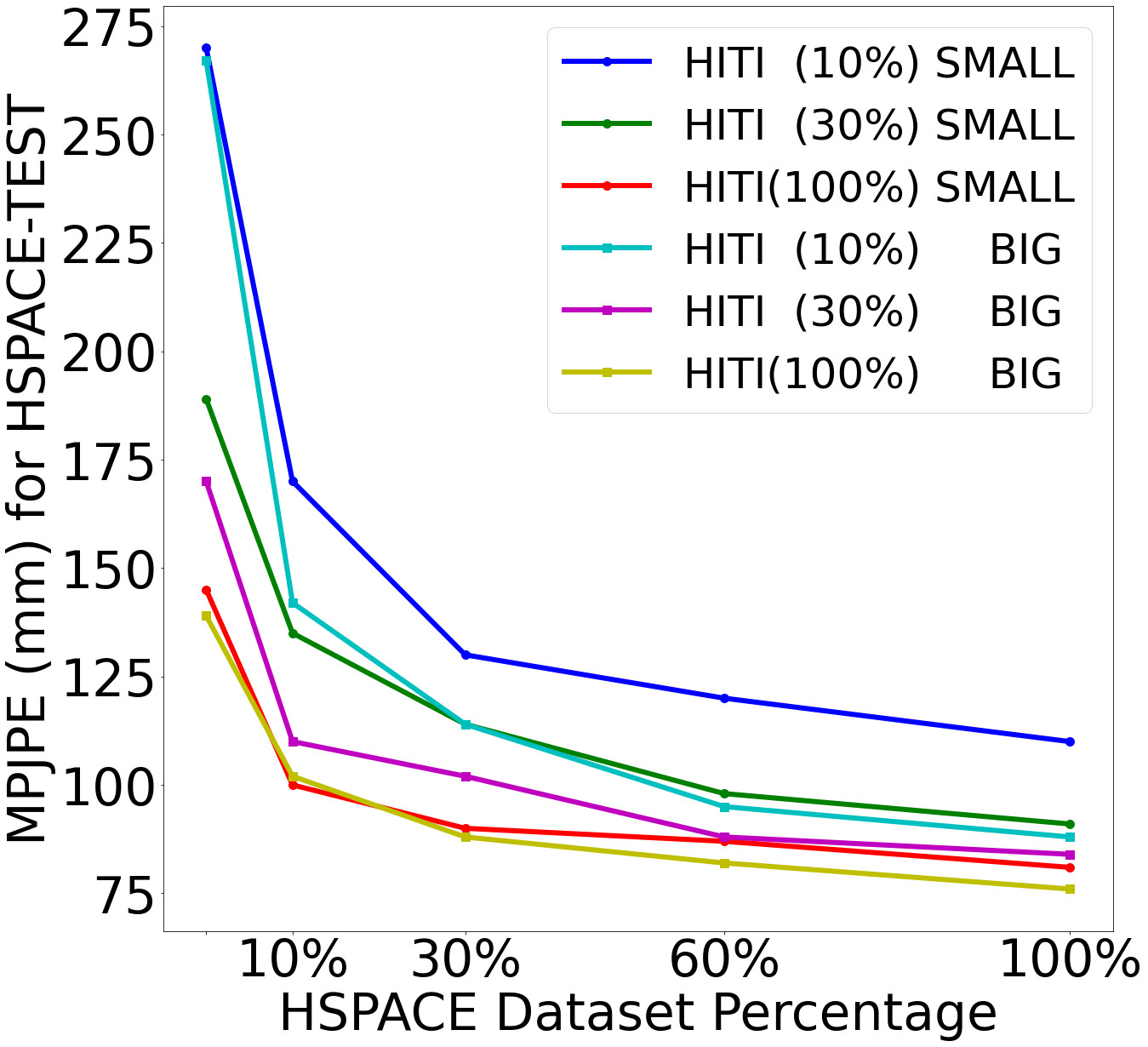}   
    \includegraphics[width=0.245\linewidth ]{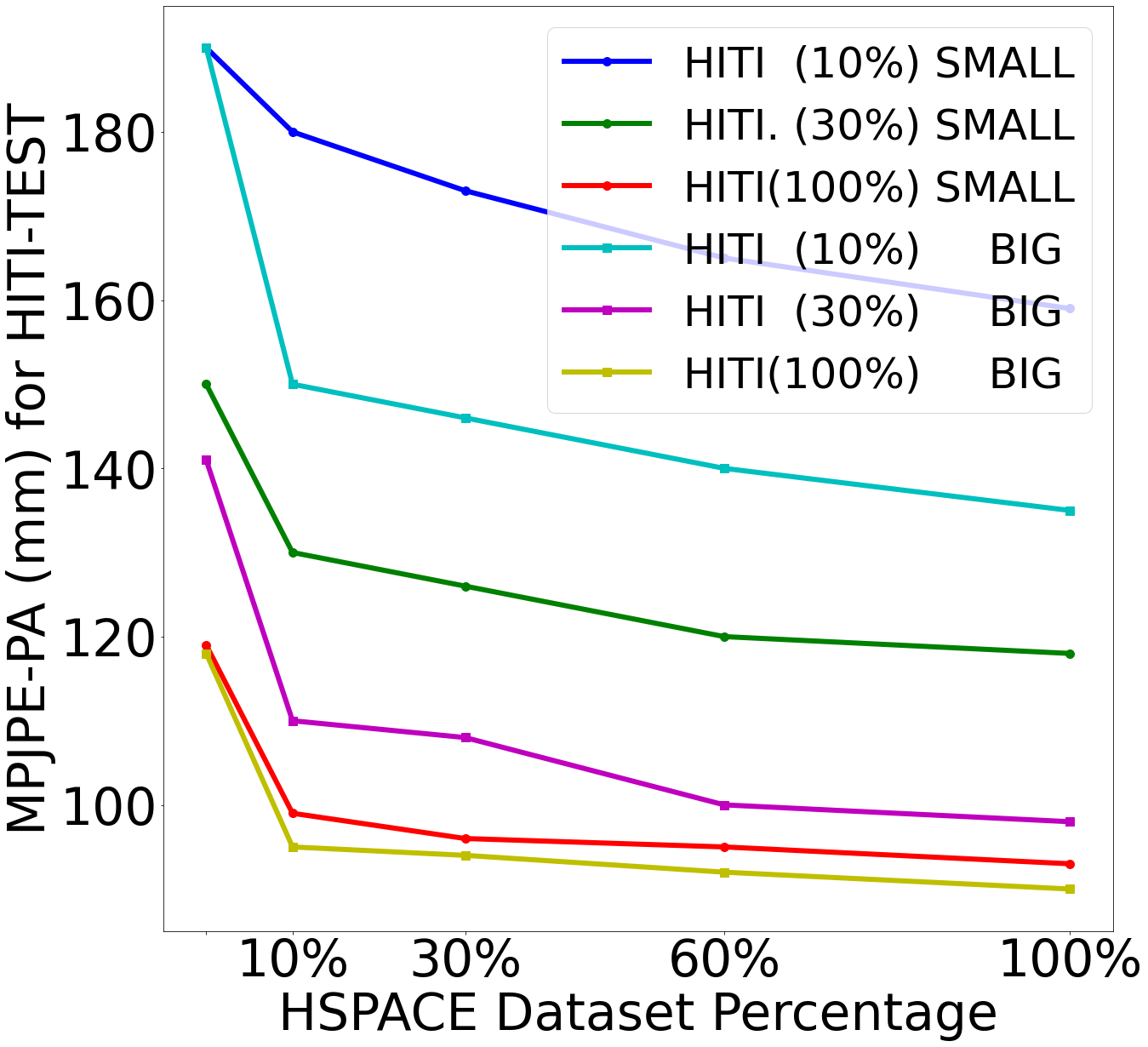}
    \includegraphics[width=0.245\linewidth ]{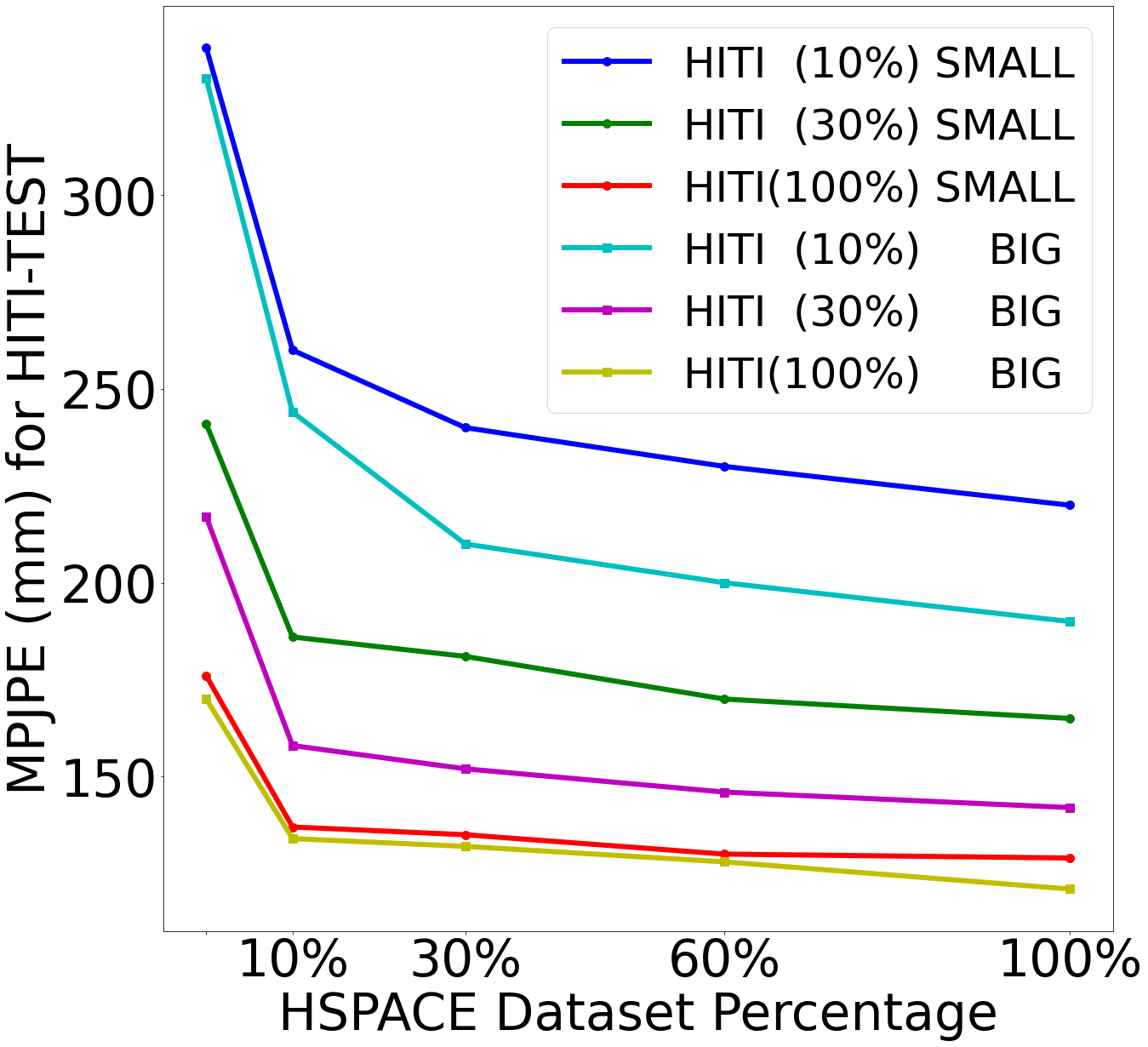}   
\end{center}
% \vspace{-4mm}
\caption{\small Performance on HSPACE-TEST set (plots in the first and second rows) and HITI-TEST set (plots in third and fourth rows) for THUNDR (WS+FS) models with different capacities (SMALL for a THUNDR model with a transformer component of 1.9M parameters and BIG for a THUNDR model with a transformer component of 3.8M parameters, see supplementary material for more details) trained with various percentages of HITI (real) and HSPACE (synthetic) data. The THUNDR models were first trained in weakly supervised (WS) regime on the percentage of HITI data indicated in the legend and then refined in a fully supervised (FS) regime on different amounts of HSPACE data as well. We report MPJPE-PA and MPJPE metrics. We observe performance improvements when adding greater amounts of both synthetic and real data, as well as when increasing the model capacity.}
\label{fig:thundr_ws_fs_ablations}
\end{figure*}

We trained a battery of models with different fractions of weakly supervised real data (10\%, 30\% or 100\% of HITI-TRAIN), fully supervised synthetic data (0\%, 10\%, 30\%, 60\%, 100\% of HSPACE-TRAIN), and for two model sizes (small THUNDR model with a transformer component of 1.9M parameters, and a big THUNDR model with a transformer component of 3.8M parameters). All models were evaluated on HSPACE-TEST (first and second columns in figure \ref{fig:thundr_ws_fs_ablations}) as well as on HITI-TEST for complex real images. Results are presented in fig. \ref{fig:thundr_ws_fs_ablations}. 

Empirically we found that models trained on synthetic data alone do not perform the best, not even when tested on synthetic data. Moreover, we found that pre-training with real data and refining on synthetic data produces better results than vice-versa. 
Large volumes of synthetic data improve model performance in conjunction with increasing amounts of weakly annotated real data, which is important as this is a practical setting and the symbiosis of synthetic and real data during training appears to address the sim-to-real gap. An increase in model capacity seems however necessary in order to take advantage of larger datasets. 

\begin{table}[!htbp]
    \small
    \centering
    \begin{tabular}[t]{|l||r|r|r|}
    \hline
    \textbf{Method}  & {MPJPE-PA} & {MPJPE} & {MPJPE-T} \\ 
    \hline
    \hline
    HMR \cite{Kanazawa2018} & $58.1$ & $88.0$& NR \\
    \hline
    HUND \cite{zanfir2020neural} & $53.0$ & $72.0$& $160.0$ \\
    \hline
    THUNDR \cite{Zanfir_2021_ICCV} & $39.8$ & $55.0$ & $143.9$ \\
    \hline
    \hline
    THUNDR (HSPACE) & $\mathbf{39.0}$ & $\mathbf{53.3}$ & $\mathbf{132.5}$ \\
    \hline
    \end{tabular}
    \caption{\small Results obtained when refining THUNDR \cite{Zanfir_2021_ICCV} on the HSPACE training set and evaluated on Human3.6M under training/testing assumptions of protocol P1 (100K testing samples). Refining on HSPACE improves over the previous SOTA under MPJPE-PA, MPJPE and translation errors (MPJPE-T).}
\label{tbl:H36MP1}
\end{table}

\begin{table*}[htbp]
    \small
    \centering
    \begin{tabular}[t]{|l||r|r|r|r|r|r|r|r|}
    \hline
    \textbf{Method}  & {MPJPE-PA} & {MPJPE} & {MPVPE-PA} & {MPVPE} & {MPJPE-T} & R\#{2D} & R\#{2D-3D} & S\#{3D} \\
    \hline
    SPIN \cite{kolotouros2019learning} & $79$ & $125$ & N/A & N/A & N/A & 111K & 300K & $0$\\
    \hline
    VIBE \cite{kocabas2019vibe} & $120$ & $260$ & N/A & N/A & N/A & 150K & 250K  & $0$ \\
    \hline
    HUND \cite{zanfir2020neural} & $84$ & $130$ & $96$ & $150$ & $280$ & 80K & 150K & $0$ \\    
    \hline
    THUNDR \cite{Zanfir_2021_ICCV} & $65$ & $100$ & $80$ & $120$ & $230$ & 80K & 150K & $0$ \\
    \hline
    \hline
    THUNDR (HITI + HSPACE) & $\mathbf{50}$ & $\textbf{76}$ & $\textbf{60}$ & $\textbf{90}$ & \textbf{180} & 100K & $0$ & $800K$ \\
    \hline
    T-THUNDR (HITI + HSPACE) & $\mathbf{47}$ & $\textbf{71}$ & $\textbf{58}$ & $\textbf{81}$ & \textbf{171} & 100K & $0$ & $800K$ \\
    \hline
    \end{tabular}

    \caption{\small Results on the \textbf{HSPACE} test set. All current state of the art methods do not perform well when tested on the HSPACE test set. However performance improves significantly when training on HSPACE. We report mean per joint positional errors (with and without Procrustes alignment) (MPJPE-PA, MPJPE), mean per joint vertex displacement error (with and without Procrustes alignment) (MPVPE-PA, MPVPE) computed against ground truth GHUM meshes and translation error (MPJPE-T) computed against the pelvis joint. We also report the number of real images and the type of annotations used during the training of the listed models, e.g. number of real images with 2d annotations (R\#2D), number of real  images with paired 2d-3d annotations (R\#2D-3D) used during training and number of synthetic images with full 3d supervision (S\#3D). See our Sup. Mat. for additional detail and for qualitative visualisations of 3d human pose and shape reconstruction.}
\label{tbl:ablation_space}
\end{table*}

\noindent{\bf Ethical Considerations.} Our dataset creation methodology aims at diversity and coverage in order to build synthetic ground-truth for different human body proportions, poses, motions, ethnicity, age, or clothing.  By generating people in new synthetic poses, and by controlling different body proportions in various scenes, we can produce considerable diversity by largely relying on synthetic assets and by varying the parameters of a statistical human pose and shape model (GHUM). This supports, in turn, our long-term goal to build inclusive models that work well for everyone especially in cases where real human data as well as forms of 3d ground truth are difficult to collect.

\section{Conclusions}

We have introduced HSPACE, a large-scale dataset of humans animated in complex synthetic indoor and outdoor environments. We combine diverse individuals of varying ages, gender, proportions, and ethnicity, with many motions and scenes, as well as parametric variations in body shape, as well as gestures, in order to generate an initial dataset of over 1 million frames. Human animations are obtained by fitting an expressive human body model, GHUM, to single scans of people, followed by re-targeting and re-posing procedures that support realistic animation, statistic variations of body proportions, and jointly consistent scene placement for multiple moving people. All assets are generated automatically, 
%at scale, 
being compatible with existing real time rendering engines. 
The dataset and an evaluation server will be made available for research.

Our quantitative evaluation of 3d human pose and shape estimation in synthetic and mixed (sim-real) regimes, underlines (1) the importance of synthetic, large-scale datasets, but also (2) the need for real data, within weakly supervised training regimes, as well as (3) the effect of increasing (matching) model capacity, for domain transfer, and continuing performance improvement as datasets grow. 
%%%%%%%%% REFERENCES

{\small
\bibliographystyle{ieee_fullname}
\bibliography{egbib}

\begin{thebibliography}{10}\itemsep=-1pt

\bibitem{renderpeople}
\url{https://renderpeople.com/}.

\bibitem{bazarevsky2020blazepose}
Valentin Bazarevsky, Ivan Grishchenko, Karthik Raveendran, Tyler Zhu, Fan
  Zhang, and Matthias Grundmann.
\newblock Blazepose: On-device real-time body pose tracking, 2020.

\bibitem{bogo2016}
Federica Bogo, Angjoo Kanazawa, Christoph Lassner, Peter Gehler, Javier Romero,
  and Michael~J Black.
\newblock Keep it {SMPL}: Automatic estimation of 3d human pose and shape from
  a single image.
\newblock In {\em ECCV}, 2016.

\bibitem{caoHMP2020}
Zhe Cao, Hang Gao, Karttikeya Mangalam, Qizhi Cai, Minh Vo, and Jitendra Malik.
\newblock Long-term human motion prediction with scene context.
\newblock In {\em ECCV}. 2020.

\bibitem{ExPose:2020}
Vasileios Choutas, Georgios Pavlakos, Timo Bolkart, Dimitrios Tzionas, and
  Michael~J. Black.
\newblock Monocular expressive body regression through body-driven attention.
\newblock In {\em European Conference on Computer Vision (ECCV)}, 2020.

\bibitem{Fieraru_2020_CVPR}
Mihai Fieraru, Mihai Zanfir, Elisabeta Oneata, Alin-Ionut Popa, Vlad Olaru, and
  Cristian Sminchisescu.
\newblock Three-dimensional reconstruction of human interactions.
\newblock In {\em Proceedings of the IEEE/CVF Conference on Computer Vision and
  Pattern Recognition (CVPR)}, June 2020.

\bibitem{fieraru2020three}
Mihai Fieraru, Mihai Zanfir, Elisabeta Oneata, Alin-Ionut Popa, Vlad Olaru, and
  Cristian Sminchisescu.
\newblock Three-dimensional reconstruction of human interactions.
\newblock In {\em Proceedings of the IEEE/CVF Conference on Computer Vision and
  Pattern Recognition}, pages 7214--7223, 2020.

\bibitem{fieraru2021learning}
Mihai Fieraru, Mihai Zanfir, Elisabeta Oneata, Alin-Ionut Popa, Vlad Olaru, and
  Cristian Sminchisescu.
\newblock Learning complex 3d human self-contact.
\newblock In {\em Thirty-Fifth AAAI Conf. on Artificial Intelligence
  (AAAI’21)}, 2021.

\bibitem{fieraru2021aifit}
Mihai Fieraru, Mihai Zanfir, Silviu~Cristian Pirlea, Vlad Olaru, and Cristian
  Sminchisescu.
\newblock Aifit: Automatic 3d human-interpretable feedback models for fitness
  training.
\newblock In {\em Proceedings of the IEEE/CVF Conference on Computer Vision and
  Pattern Recognition}, 2021.

\bibitem{fieraru2021remips}
Mihai Fieraru, Mihai Zanfir, Teodor~Alexandru Szente, Eduard~Gabriel Bazavan,
  Vlad Olaru, and Cristian Sminchisescu.
\newblock {REMIPS}: Physically consistent 3d reconstruction of multiple
  interacting people under weak supervision.
\newblock In {\em Thirty-Fifth Conference on Neural Information Processing
  Systems}, 2021.

\bibitem{Hansen2006}
Nikolaus Hansen.
\newblock {\em The CMA Evolution Strategy: A Comparing Review}, pages 75--102.
\newblock Springer Berlin Heidelberg, Berlin, Heidelberg, 2006.

\bibitem{huang2020arch}
Zeng Huang, Yuanlu Xu, Christoph Lassner, Hao Li, and Tony Tung.
\newblock Arch: Animatable reconstruction of clothed humans.
\newblock In {\em Proceedings of the IEEE/CVF Conference on Computer Vision and
  Pattern Recognition}, pages 3093--3102, 2020.

\bibitem{Ionescu14pami}
C. Ionescu, D. Papava, V. Olaru, and C. Sminchisescu.
\newblock Human3.6{M}: Large scale datasets and predictive methods for 3d human
  sensing in natural environments.
\newblock {\em PAMI}, 2014.

\bibitem{Jacobson:WN:2013}
Alec Jacobson, Ladislav Kavan, and Olga Sorkine-Hornung.
\newblock Robust inside-outside segmentation using generalized winding numbers.
\newblock {\em ACM Transactions on Graphics (proceedings of ACM SIGGRAPH)},
  32(4):33:1--33:12, 2013.

\bibitem{jiang2020coherent}
Wen Jiang, Nikos Kolotouros, Georgios Pavlakos, Xiaowei Zhou, and Kostas
  Daniilidis.
\newblock Coherent reconstruction of multiple humans from a single image.
\newblock In {\em CVPR}, pages 5579--5588, 2020.

\bibitem{Joo_2015_ICCV}
Hanbyul Joo, Hao Liu, Lei Tan, Lin Gui, Bart Nabbe, Iain Matthews, Takeo
  Kanade, Shohei Nobuhara, and Yaser Sheikh.
\newblock Panoptic studio: A massively multiview system for social motion
  capture.
\newblock In {\em ICCV}, 2015.

\bibitem{joo2020exemplar}
Hanbyul Joo, Natalia Neverova, and Andrea Vedaldi.
\newblock Exemplar fine-tuning for 3d human pose fitting towards in-the-wild 3d
  human pose estimation.
\newblock {\em arXiv preprint arXiv:2004.03686}, 2020.

\bibitem{Kanazawa2018}
Angjoo Kanazawa, Michael~J. Black, David~W. Jacobs, and Jitendra Malik.
\newblock End-to-end recovery of human shape and pose.
\newblock In {\em CVPR}, 2018.

\bibitem{kocabas2019vibe}
Muhammed Kocabas, Nikos Athanasiou, and Michael~J Black.
\newblock Vibe: Video inference for human body pose and shape estimation.
\newblock {\em CVPR}, 2020.

\bibitem{kolotouros2019learning}
Nikos Kolotouros, Georgios Pavlakos, Michael~J Black, and Kostas Daniilidis.
\newblock Learning to reconstruct 3d human pose and shape via model-fitting in
  the loop.
\newblock In {\em Proceedings of the IEEE International Conference on Computer
  Vision}, pages 2252--2261, 2019.

\bibitem{OpenImages}
Alina Kuznetsova, Hassan Rom, Neil Alldrin, Jasper Uijlings, Ivan Krasin, Jordi
  Pont-Tuset, Shahab Kamali, Stefan Popov, Matteo Malloci, Tom Duerig, and
  Vittorio Ferrari.
\newblock The open images dataset v4: Unified image classification, object
  detection, and visual relationship detection at scale.
\newblock {\em arXiv:1811.00982}, 2018.

\bibitem{MsCOCO}
Tsung{-}Yi Lin, Michael Maire, Serge~J. Belongie, Lubomir~D. Bourdev, Ross~B.
  Girshick, James Hays, Pietro Perona, Deva Ramanan, Piotr Doll{\'{a}}r, and
  C.~Lawrence Zitnick.
\newblock Microsoft {COCO:} common objects in context.
\newblock {\em CoRR}, abs/1405.0312, 2014.

\bibitem{SMPL2015}
Matthew Loper, Naureen Mahmood, Javier Romero, Gerard Pons-Moll, and Michael~J.
  Black.
\newblock {SMPL}: A skinned multi-person linear model.
\newblock {\em SIGGRAPH}, 2015.

\bibitem{mehta2017monocular}
Dushyant Mehta, Helge Rhodin, Dan Casas, Pascal Fua, Oleksandr Sotnychenko,
  Weipeng Xu, and Christian Theobalt.
\newblock Monocular 3d human pose estimation in the wild using improved cnn
  supervision.
\newblock In {\em 2017 international conference on 3D vision (3DV)}, pages
  506--516. IEEE, 2017.

\bibitem{singleshotmultiperson2018}
Dushyant Mehta, Oleksandr Sotnychenko, Franziska Mueller, Weipeng Xu, Srinath
  Sridhar, Gerard Pons-Moll, and Christian Theobalt.
\newblock Single-shot multi-person 3d pose estimation from monocular rgb.
\newblock In {\em 3D Vision (3DV), 2018 Sixth International Conference on}.
  IEEE, sep 2018.

\bibitem{Mueller:CVPR:2021}
Lea M{\"u}ller, Ahmed A.~A. Osman, Siyu Tang, Chun-Hao~P. Huang, and Michael~J.
  Black.
\newblock On self-contact and human pose.
\newblock In {\em Proceedings IEEE/CVF Conf.~on Computer Vision and Pattern
  Recognition (CVPR)}, June 2021.

\bibitem{Patel:CVPR:2021}
Priyanka Patel, Chun-Hao~P. Huang, Joachim Tesch, David~T. Hoffmann, Shashank
  Tripathi, and Michael~J. Black.
\newblock {AGORA}: Avatars in geography optimized for regression analysis.
\newblock In {\em Proceedings IEEE/CVF Conf.~on Computer Vision and Pattern
  Recognition ({CVPR})}, June 2021.

\bibitem{pavlakoscvpr2019}
Georgios Pavlakos, Vasileios Choutas, Nima Ghorbani, Timo Bolkart, Ahmed Osman,
  Dimitrios Tzionas, and Michael Black.
\newblock Expressive body capture: 3d hands, face, and body from a single
  image.
\newblock In {\em CVPR}, 2019.

\bibitem{dmhs_cvpr17}
A.I. Popa, M. Zanfir, and C. Sminchisescu.
\newblock {Deep Multitask Architecture for Integrated 2D and 3D Human Sensing}.
\newblock In {\em CVPR}, 2017.

\bibitem{premecz2006iterative}
M{\'a}ty{\'a}s Premecz.
\newblock Iterative parallax mapping with slope information.
\newblock In {\em Central European Seminar on Computer Graphics}, volume~1,
  pages 1--8. Citeseer, 2006.

\bibitem{pumarola20193dpeople}
Albert Pumarola, Jordi Sanchez-Riera, Gary Choi, Alberto Sanfeliu, and Francesc
  Moreno-Noguer.
\newblock 3dpeople: Modeling the geometry of dressed humans.
\newblock In {\em Proceedings of the IEEE/CVF International Conference on
  Computer Vision}, pages 2242--2251, 2019.

\bibitem{Renderpeople.com}
Renderpeople.
\newblock Renderpeople: 3d people for renderings.

\bibitem{Rhodin_2018_ECCV}
Helge Rhodin, Mathieu Salzmann, and Pascal Fua.
\newblock Unsupervised geometry-aware representation for 3d human pose
  estimation.
\newblock In {\em ECCV}, September 2018.

\bibitem{Guler2018DensePose}
Iasonas~Kokkinos Riza Alp~Guler, Natalia~Neverova.
\newblock Densepose: Dense human pose estimation in the wild.
\newblock {\em arXiv}, 2018.

\bibitem{STRAPS2020BMVC}
Akash Sengupta, Ignas Budvytis, and Roberto Cipolla.
\newblock Synthetic training for accurate 3d human pose and shape estimation in
  the wild.
\newblock In {\em British Machine Vision Conference (BMVC)}, September 2020.

\bibitem{sigal2010humaneva}
Leonid Sigal, Alexandru~O Balan, and Michael~J Black.
\newblock Humaneva: Synchronized video and motion capture dataset and baseline
  algorithm for evaluation of articulated human motion.
\newblock {\em International journal of computer vision}, 87(1-2):4, 2010.

\bibitem{sminchisescu_ijrr03}
C. Sminchisescu and B. Triggs.
\newblock {Estimating Articulated Human Motion with Covariance Scaled
  Sampling}.
\newblock {\em IJRR}, 22(6):371--393, 2003.

\bibitem{trumble2017total}
Matthew Trumble, Andrew Gilbert, Charles Malleson, Adrian Hilton, and John~P
  Collomosse.
\newblock Total capture: 3d human pose estimation fusing video and inertial
  sensors.
\newblock In {\em BMVC}, volume~2, pages 1--13, 2017.

\bibitem{varol2017learning}
Gul Varol, Javier Romero, Xavier Martin, Naureen Mahmood, Michael~J Black, Ivan
  Laptev, and Cordelia Schmid.
\newblock Learning from synthetic humans.
\newblock In {\em Proceedings of the IEEE Conference on Computer Vision and
  Pattern Recognition}, pages 109--117, 2017.

\bibitem{vonMarcard2018}
Timo von Marcard, Roberto Henschel, Michael Black, Bodo Rosenhahn, and Gerard
  Pons-Moll.
\newblock Recovering accurate 3d human pose in the wild using {IMUs} and a
  moving camera.
\newblock In {\em ECCV}, 2018.

\bibitem{ghum2020}
Hongyi Xu, Eduard~Gabriel Bazavan, Andrei Zanfir, Bill Freeman, Rahul
  Sukthankar, and Cristian Sminchisescu.
\newblock {GHUM} \& {GHUML}: Generative {3D} human shape and articulated pose
  models.
\newblock {\em CVPR}, 2020.

\bibitem{yan2021ultrapose}
Haonan Yan, Jiaqi Chen, Xujie Zhang, Shengkai Zhang, Nianhong Jiao, Xiaodan
  Liang, and Tianxiang Zheng.
\newblock Ultrapose: Synthesizing dense pose with 1 billion points by
  human-body decoupling 3d model.
\newblock In {\em Proceedings of the IEEE/CVF International Conference on
  Computer Vision}, pages 10891--10900, 2021.

\bibitem{yu2020humbi}
Zhixuan Yu, Jae~Shin Yoon, In~Kyu Lee, Prashanth Venkatesh, Jaesik Park, Jihun
  Yu, and Hyun~Soo Park.
\newblock Humbi: A large multiview dataset of human body expressions.
\newblock In {\em Proceedings of the IEEE/CVF Conference on Computer Vision and
  Pattern Recognition}, pages 2990--3000, 2020.

\bibitem{zanfir2020weakly}
Andrei Zanfir, Eduard~Gabriel Bazavan, Hongyi Xu, Bill Freeman, Rahul
  Sukthankar, and Cristian Sminchisescu.
\newblock Weakly supervised 3d human pose and shape reconstruction with
  normalizing flows.
\newblock {\em ECCV}, 2020.

\bibitem{zanfir2020neural}
Andrei Zanfir, Eduard~Gabriel Bazavan, Mihai Zanfir, William~T Freeman, Rahul
  Sukthankar, and Cristian Sminchisescu.
\newblock Neural descent for visual 3d human pose and shape.
\newblock {\em arXiv preprint arXiv:2008.06910}, 2020.

\bibitem{zanfir2018monocular}
Andrei Zanfir, Elisabeta Marinoiu, and Cristian Sminchisescu.
\newblock Monocular 3d pose and shape estimation of multiple people in natural
  scenes-the importance of multiple scene constraints.
\newblock In {\em CVPR}, 2018.

\bibitem{Zanfir_2021_ICCV}
Mihai Zanfir, Andrei Zanfir, Eduard~Gabriel Bazavan, William~T. Freeman, Rahul
  Sukthankar, and Cristian Sminchisescu.
\newblock Thundr: Transformer-based 3d human reconstruction with markers.
\newblock In {\em Proceedings of the IEEE/CVF International Conference on
  Computer Vision (ICCV)}, pages 12971--12980, October 2021.

\bibitem{zhang2021body}
Jianfeng Zhang, Dongdong Yu, Jun~Hao Liew, Xuecheng Nie, and Jiashi Feng.
\newblock Body meshes as points.
\newblock In {\em Proceedings of the IEEE/CVF Conference on Computer Vision and
  Pattern Recognition}, pages 546--556, 2021.

\bibitem{zhang2020object}
Tianshu Zhang, Buzhen Huang, and Yangang Wang.
\newblock Object-occluded human shape and pose estimation from a single color
  image.
\newblock In {\em Proceedings of the IEEE/CVF Conference on Computer Vision and
  Pattern Recognition}, pages 7376--7385, 2020.

\bibitem{zhou2018continuity}
Yi Zhou, Connelly Barnes, Jingwan Lu, Jimei Yang, and Hao Li.
\newblock On the continuity of rotation representations in neural networks.
\newblock {\em arXiv preprint arXiv:1812.07035}, 2018.

\bibitem{zhu2020simpose}
Tyler Zhu, Per Karlsson, and Christoph Bregler.
\newblock Simpose: Effectively learning densepose and surface normals of people
  from simulated data, 2020.

\end{thebibliography}
}

\end{document}